\pdfoutput=1
\documentclass{article}

\usepackage{microtype}
\usepackage{graphicx}
\usepackage{booktabs} 
\usepackage[T1]{fontenc}

\usepackage{hyperref}

\usepackage{amssymb}
\usepackage{microtype}
\usepackage{graphicx}
\usepackage{subcaption}

\usepackage{amsfonts}
\usepackage{amsthm}
\usepackage{amsmath}
\usepackage{soul}
\usepackage{algorithm2e}

\DeclareMathOperator*{\argmax}{arg\,max}

\newcommand{\fig}[1]{Figure~\ref{fig:#1}}

\newcommand{\alg}[1]{Algorithm~\ref{alg:#1}}
\newcommand{\eq}[1]{(\ref{eq:#1})}

\usepackage{enumitem}

\usepackage[accepted]{icml2020}

\icmltitlerunning{Towards Similarity Graphs Constructed by Deep Reinforcement Learning}

\begin{document}

\twocolumn[
\icmltitle{Towards Similarity Graphs Constructed \\by Deep Reinforcement Learning}



\icmlsetsymbol{equal}{*}

\begin{icmlauthorlist}
\icmlauthor{Dmitry Baranchuk}{yandex,msu}
\icmlauthor{Artem Babenko}{yandex,hse}
\end{icmlauthorlist}

\icmlaffiliation{yandex}{Yandex, Russia}
\icmlaffiliation{msu}{Lomonosov Moscow State University, Russia}
\icmlaffiliation{hse}{National Research University Higher School of Economics, Russia}

\icmlcorrespondingauthor{Dmitry Baranchuk}{dmitry.baranchuk@graphics.cs.msu.ru}

\vskip 0.3in
]
\printAffiliationsAndNotice{}  

\begin{abstract}
Similarity graphs are an active research direction for the nearest neighbor search (NNS) problem. New algorithms for similarity graph construction are continuously being proposed and analyzed by both theoreticians and practitioners. However, existing construction algorithms are mostly based on heuristics and do not explicitly maximize the target performance measure, i.e., search recall. Therefore, at the moment it is not clear whether the performance of similarity graphs has plateaued or more effective graphs can be constructed with more theoretically grounded methods. In this paper, we introduce a new principled algorithm, based on adjacency matrix optimization, which explicitly maximizes search efficiency. Namely, we propose a probabilistic model of a similarity graph defined in terms of its edge probabilities and show how to learn these probabilities from data as a reinforcement learning task. As confirmed by experiments, the proposed construction method can be used to refine the state-of-the-art similarity graphs, achieving higher recall rates for the same number of distance computations. Furthermore, we analyze the learned graphs and reveal the structural properties that are responsible for more efficient search.

\end{abstract}

\section{Introduction}
\label{sect:intro}

In this paper, we address nearest neighbor search (NNS), a long-standing problem, arising in a large number of machine learning applications, such as recommender services, information retrieval, and others. The NNS problem is formalized as follows. Given the database $D = \{v_1,\dots,v_N\} \subset \mathbb{R}^d$ and a query $q \in \mathbb{R}^d$, one needs to find the datapoint $v \in D$ that is closest to the query in terms of some distance (e.g. Euclidean). As the sizes of databases $|D|$ in arising practical tasks are constantly increasing, the efficiency and the scalability of NNS become crucial. 

Thus, the problem of efficient NNS receives much attention from the machine learning community. Well-known established approaches, based on partition trees \cite{KdTree,PcaTree,ApdTree,RpTree,dasgupta2013randomized} and locality-sensitive hashing (LSH) \cite{LSH98,LSH,andoni2008near,Razenshteyn15} have been developed by ML researchers for decades and provide both decent practical performance and theoretical guarantees. Recently, similarity graph methods \cite{navarro2002searching,malkov2018efficient,fu2016efanna,NSG}, were shown to outperform tree-based and LSH-based techniques \cite{annbenchmarks}. These methods represent the database as a graph, and at the search stage, a query traverses the graph via beam search. While these methods do not have full theoretical support yet, their exceptional practical performance has shifted the research attention to the development of new approaches based on this paradigm.

Due to the great importance of the NNS problem, new algorithms for similarity graphs construction are being proposed and analyzed by both theoreticians \cite{Laarhoven18} and practitioners \cite{fu2016efanna,malkov2018efficient,NSG,iwasaki2018optimization}. Most of these works, however, propose new heuristics-based procedures, which do not explicitly optimize search efficiency. Moreover, different methods often achieve superior results only on a subset of datasets, which implies that the proposed heuristics are not universally applicable.

In this work, we introduce a new method for similarity graph construction that explicitly maximizes the search efficiency via optimization of the graph adjacency matrix. Specifically, we define a probabilistic model of a similarity graph in terms of its edge probabilities. Then we learn these probabilities from data, maximizing the search efficiency for a large set of training queries. It appears that this task could be naturally treated as a reinforcement learning problem. As a result, the proposed algorithm produces a graph that outperforms graphs constructed via heuristic approaches. 

To sum up, the contributions of this paper are as follows:

\begin{enumerate}
    \vspace{-1px}
    \item We develop a new algorithm for similarity graph construction that explicitly optimizes search efficiency. To the best of our knowledge, all existing methods are based on heuristics that can have limited niches of applicability.
    \item By experiments on common benchmarks, we show that the proposed algorithm can be used to refine state-of-the-art similarity graphs, which allows to achieve higher recall rates under the same number of distance computations. We also analyse the learned graphs and investigate the properties that cause the gains.
    \item We demonstrate a novel practical large-scale application of the reinforcement learning machinery that explicitly optimizes the quality of similarity graphs with millions of edges.
    \vspace{-1px}
\end{enumerate}

The rest of the paper is organized as follows. First, we discuss relevant prior works. Then we describe the proposed RL-based graph construction algorithm, empirically analyze it and confirm its advantage over heuristic-based methods. The source code of our algorithm and experiments are available online\footnote[1]{https://github.com/dbaranchuk/nns-meets-deep-rl}

\section{Related work}
\label{sect:related}

Here we briefly review the ideas from the prior works that are relevant to our approach.

\textbf{Nearest neighbor search techniques.} The existing NNS approaches mostly fall into three research directions. Methods from the first direction, based on partition trees  \cite{KdTree,PcaTree,ApdTree,RpTree,dasgupta2013randomized}, hierarchically split the search space into a large number of regions, corresponding to tree leaves, and the query visits only a limited number of promising regions when searching. Second, locality-sensitive hashing methods \cite{LSH98,LSH,andoni2008near,Razenshteyn15} map the database points into several buckets using several hash functions such that the probability of collision is much higher for nearby points than for points that are further apart. At the search stage, a query is also hashed, and distances to all the points from the corresponding buckets are evaluated. The third direction of similarity graphs   \cite{navarro2002searching,malkov2018efficient,fu2016efanna,NSG,iwasaki2018optimization} represents the database as a directed graph, and on the search stage, a query traverses the graph via beam search. The empirical performance of similarity graphs was shown to be much higher compared to LSH-based and tree-based methods \cite{HNSW}. In more details, the typical search process in similarity graphs performs as follows. The database is organized in a graph, where each vertex corresponds to some datapoint, and the vertices, corresponding to the neighboring datapoints, are connected by edges. The search algorithm picks a start vertex (random or predefined) and iteratively explores the graph from it. On each iteration, the query tries to improve its position by moving to a vertex from a candidate pool that is closest to the query. The routing process stops when there are no closer vertices in the pool.

\textbf{Similarity graphs construction procedures.} Several recent works developing similarity graph methods typically differ in graph construction procedures, based on different heuristics. For instance, the recent HNSW algorithm \cite{HNSW} performs consecutive insertions of database items into the graph structure. This procedure provides long-range edges for efficient graph navigation. Moreover, an additional structure of a nested hierarchy of layers is proposed for further speedup. Another recent graph, NSG \cite{NSG}, employs a k-nearest neighbor graph as an initial graph structure, then performs the search procedure with each node being a query, connects the node with vertices visited during the search and selects edges following the pruning strategy. The recently proposed graph-based method NGT-onng \cite{iwasaki2018optimization} provides a set of heuristics for graph construction and finds optimal indegrees and outdegrees for a specific precision region.

Both \cite{NSG,iwasaki2018optimization} report that the advantage of different graphs is revealed on different datasets, which implies the limitations of the heuristics in use. Instead, our approach aims to learn the graph from data, explicitly optimizing the search efficiency.

\textbf{Learning of data structures.}
The recent line of works \cite{kraska2018case,kraska2019sagedb} proposes to use machine learning methods instead of the traditional database indices, such as B-trees and Bloom Filters. While being related, these methods are not directly applied to the construction of similarity graphs, which we address in this paper.

\textbf{Reinforcement learning for discrete structures.} Our approach is partially inspired by the recent RL success for structure learning in different machine learning pipelines. Probably, the most well-known use-case is the learning of DNN structure \cite{zoph2016neural}. Another related recent work is DeepPath \cite{xiong2017deeppath} that employs RL to learn structures of the knowledge graphs. In this paper, we demonstrate that RL is also a natural fit for the problem of similarity graph construction for NNS. 


\section{Method}
\label{sect:method}

In this section, we describe our approach for similarity graph construction based on reinforcement learning.

\subsection{Similarity graph construction as an optimization problem}

First, we introduce a probabilistic model of a similarity graph. Our model defines a probability of a graph as a joint probability of individual edges. Each edge is modelled as an independent Bernoulli random variable $b_i \sim Bern(p_i)$ that determines whether this edge should exist in the graph. Therefore, the probability of the graph $G$ is a product of probabilities of all edges: $P(G) = P(b_1, b_2, ..., b_n) = \prod_i p^{b_i}_i(1-p_i)^{1-b_i}$. Our goal then is to maximize the following objective:
\vspace{1mm}
\begin{equation} \label{eq:optimization}
\begin{gathered}
    P^*(G){=}\argmax_{P(G)} E_{q\sim p(q)}E_{G\sim P(G)}\mathcal{R}(G, q) 
    \\
    \mathcal{R}(G, q)=\mathcal{F}(Accuracy(G, q), Complexity(G,q))
  \end{gathered}
\end{equation}

Here $E_{q\sim p(q)}$ denotes the expectation over the query distribution. $Accuracy(G, q)$ and $Complexity(G,q)$ are responsible for high search recall and high search efficiency respectively. $\mathcal{F}(\cdot,\cdot)$ plays a role of an "acquisition" function that combines both $Accuracy(G, q)$ and $Complexity(G,q)$ into one scalar value. We elaborate on each of these terms in the next section.

By solving the optimization problem \eq{optimization}, we find the edge probabilities $\{p_1,\dots,p_n\}$ that maximize the accuracy and minimize the search complexity in expectation over graphs $G \sim P(G)$. 

Finally, we obtain a deterministic graph\footnote[2]{Here we exploit the fact that our optimization problem (1) has a deterministic solution i.e. a graph where $p \in \{0, 1\}$. This property holds because our problem is equivalent to a Markov Decision Process. It can be proven that all MDPs have a deterministic optimal policy \cite{MDP}.} as $G^*{=}\argmax_{G}P^*(G)$, which corresponds to keeping the edges with $p \ge 0.5$ and omitting the edges with $p < 0.5$. This graph then can be used for NNS with one of the standard search algorithms. 

For large-scale problems, optimizing over a quadratic number of edges is infeasible. In this case we take some initial similarity graph $\hat G$ and refine it, pruning its edges via optimization \eq{optimization} over edges presented in $\hat G$. We obtain a subgraph $G^* \subseteq \hat G$ that is more efficient in terms of nearest neighbor search performance. For small-scale datasets, we aim to optimize the complete graph since it is guaranteed to contain the optimal one.

\subsection{Markov Decision Process}

Now let us formulate the optimization problem \eq{optimization} as a Markov Decision Process (MDP). We consider the initial graph $\hat G$ and search algorithm as the environment $\mathcal{E}$. An MDP agent interacts with the environment using two available actions $a$: "remove" or "keep" an edge. The environment state $s = (q, v_i, v_{adj}, V, H)$ consists of a query $q$, current vertex $v_i$, its adjacent vertices $v_{adj}$, already visited vertices $V$ and a heap of candidates $H$. The transition function $\mathcal{T}$ represents the search algorithm. In our work we exploit the standard HNSW search algorithm \cite{HNSW} and incorporate the RL agent in the loop, see \alg{beam_search}.

\vspace{1.5mm}
\begin{algorithm}[H]
\SetAlgoNoLine
 \caption{The nearest neighbor search algorithm with incorporated RL agent. \vspace{1.5mm}}

 \KwData{graph $\hat G$, query $q$, initial vertex $v_0$, output size $k$}
 
 \vspace{0.7mm}
\textbf{Initialization:}
 
 $V \xleftarrow{} \{v_0\}$  // a set of visited vertices
 
 $H \xleftarrow{} \{v_0 : d(v_0, q)\}$  // a heap of candidates
 
 $TopK \xleftarrow{} \{v_0 : d(v_0, q)\}$  // a heap of top-k results
 \vspace{0.7mm}
 
 \While{not \text{should\_stop} }{
    \vspace{0.7mm}
    /* $i$-th search step */ 
    
    $v_i \xleftarrow{}$ \text{extract nearest element from $H$ to $q$}
    
    $v_{adj} \xleftarrow{}$ \text{get adjacent vertices of $v_i$}
    
    
    $s \xleftarrow{} (q, v_i, v_{adj}, V, H)$    // collect environment state
    
    $\hat v_{adj} \xleftarrow{}$ Agent($s$) // predict what connections to keep
    
    \vspace{0.7mm}
    \For{$\hat v \in \hat v_{adj} \setminus V$}{
        
        $V \xleftarrow{} Add(V, \hat v)$
        
        $H \xleftarrow{} Insert(H, \hat v, d(\hat v, q))$
        
        $TopK \xleftarrow{} Update(TopK, k, \hat v, d(\hat v, q))$
        
    }
}

\vspace{0.7mm}
\Return $TopK$

\label{alg:beam_search}
\end{algorithm}

\textbf{Sessions}. We introduce a session $\tau$ as a search procedure for a single query $q$. On each step, the search procedure visits a vertex and updates the state $s$. The agent obtains $s$ and decides which edges are available from that vertex. In turn, the search algorithm processes the kept edges and picks the next vertex. After the search terminates, the agent obtains a reward $\mathcal{R}$ for the entire session.

\textbf{Reward function}.
Our reward function $\mathcal{R}(\tau)$ combines two components: accuracy and complexity of the search process. The accuracy for one session is an indicator $I[\tau]$ if the actual nearest neighbor is found. This term encourages the agent to maximize search recall. For instance, it may exclude edges that cause the search procedure to get stuck in poor local optima.
The second component measures the search complexity as a number of distance computations $DCS$ during one session. This term effectively encourages the agent to prune irrelevant edges.

We define the reward function as:
\begin{equation} \label{eq:reward}
\mathcal{R}(\tau) = I[\tau] \cdot \max{(DCS_{max} - DCS, 1)}
\end{equation}

where $DCS_{max}$ is a distance computation budget, which is set to restrict the search complexity for each query. Intuitively, we want the agent to find the actual nearest neighbor and then to reduce the complexity without an accuracy drop. 
If the nearest neighbor is not found then $R(\tau){=}0$ regardless of $DCS$, otherwise the agent obtains higher reward for more computationally efficient sessions. With lower $DCS_{max}$ values, the agent is more prone to sacrificing accuracy on some queries for more efficient search on others. We also observe that the value of $DCS_{max}$ affects the algorithm convergence by changing the "sharpness" of the objective function. In practice, we tune this parameter empirically based on average vertex degree and the desired recall region.

\subsection{Policy Network Architecture}
In our method, the agent is a policy network that predicts edge probabilities. For simplicity, we use a feed-forward architecture that processes each edge individually: $\pi_\theta(b|s){=}\prod^n_i{\pi_\theta(b_i|x_i(s))}$. The network receives an edge, represented as a concatenation of source and target vertices $x_i(s){=}\left[v_{source}, v_{target}\right]$, as input and predicts its probability. The network itself consists of two linear layers with ELU activations followed by another linear layer with sigmoid non-linearity. While more powerful network architectures can be used (e.g., Graph Convolutional Networks \cite{kipf2016semi}), they are typically inapplicable in the large-scale scenario due to GPU memory constraints and long training time.

\subsection{Policy optimization}
\begin{figure}
\noindent
\centering
\renewcommand\arraystretch{0.8}
\hspace{-2mm}
\includegraphics[height=5.3cm]{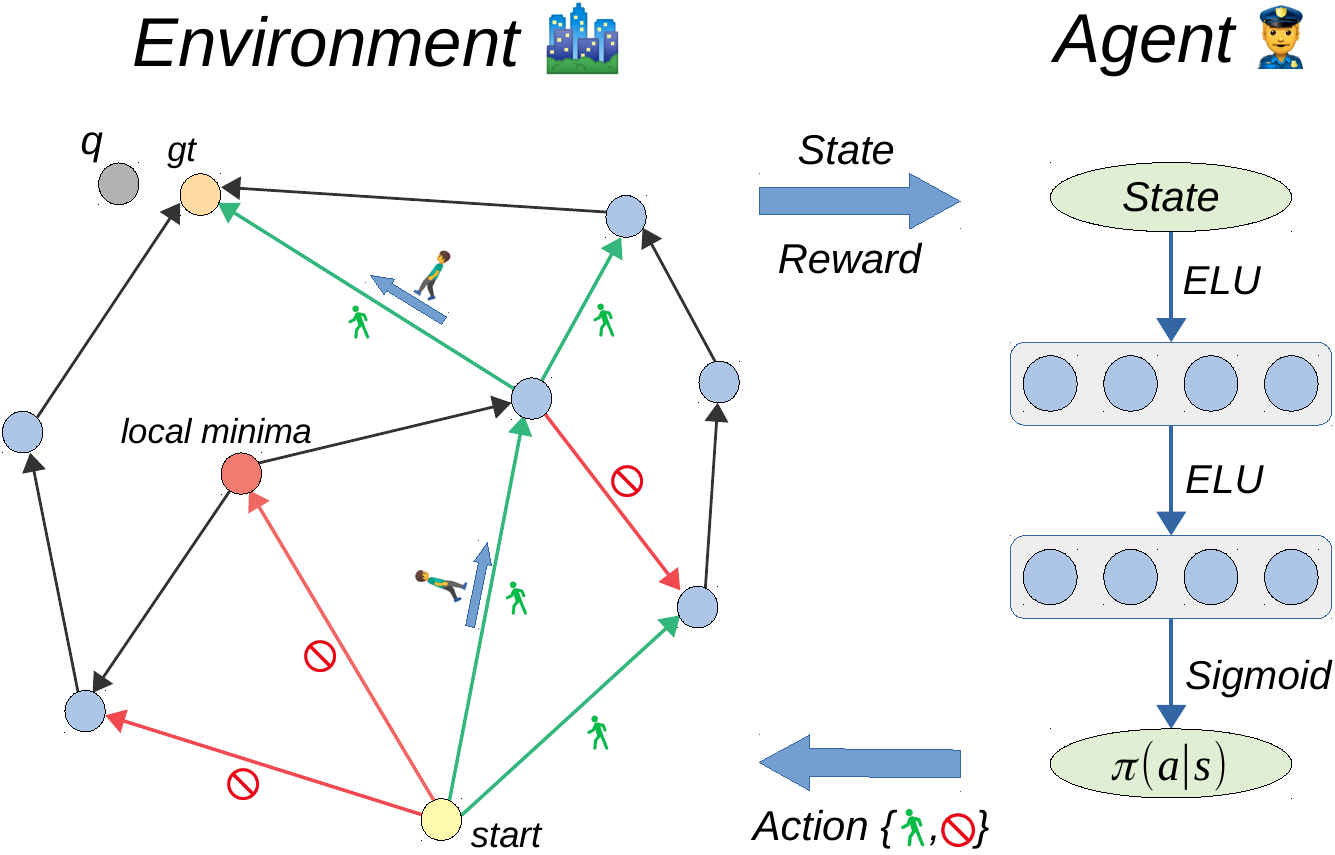}

\caption{Overview of the proposed RL scheme for graph construction. It is presented as a communication between the environment and agent. \textbf{Left:} the environment is a similarity graph equipped with a search algorithm. On each step, the search algorithm visits a node and updates the environment state. \textbf{Right:} the agent obtains the state and uses policy network to predict which outgoing edges to preserve. Then, the search procedure processes the kept edges and transits to the next node. When the search terminates, the agent obtains a total reward for the entire session. }
\label{fig:env}
\vspace{-3mm}
\end{figure}

We can now apply policy-based RL to directly optimize the expected reward \eq{reward}. The overall scheme of our approach is presented in \fig{env}. 

Among policy-based methods such as REINFORCE \cite{reinforce}, PPO \cite{ppo}, ACKTR \cite{acktr}, etc, we have found that TRPO \cite{trpo} provides the fastest convergence and the highest reward values. The main practical drawback of TRPO is that it requires a large number of sessions to perform an accurate natural gradient update. However, in our case, each session requires only a single run of the search algorithm, hence we can efficiently sample a large number of search trajectories in parallel.

We also adapt two common policy optimization tricks for our setting. First, we use reward baselines to speed up convergence by reducing gradient variance. Our algorithm maintains an individual baseline for each training query as a moving average of observed rewards for that query. Second, we facilitate exploration by adding policy entropy to the training objective. This long-standing technique \cite{reinforce} discourages the agent from premature convergence to a suboptimal deterministic policy.

\begin{figure*}
\noindent
\centering
\renewcommand\arraystretch{0.8}
\begin{tabular}{cc} \hspace{-5mm}
\includegraphics[height=8.3cm]{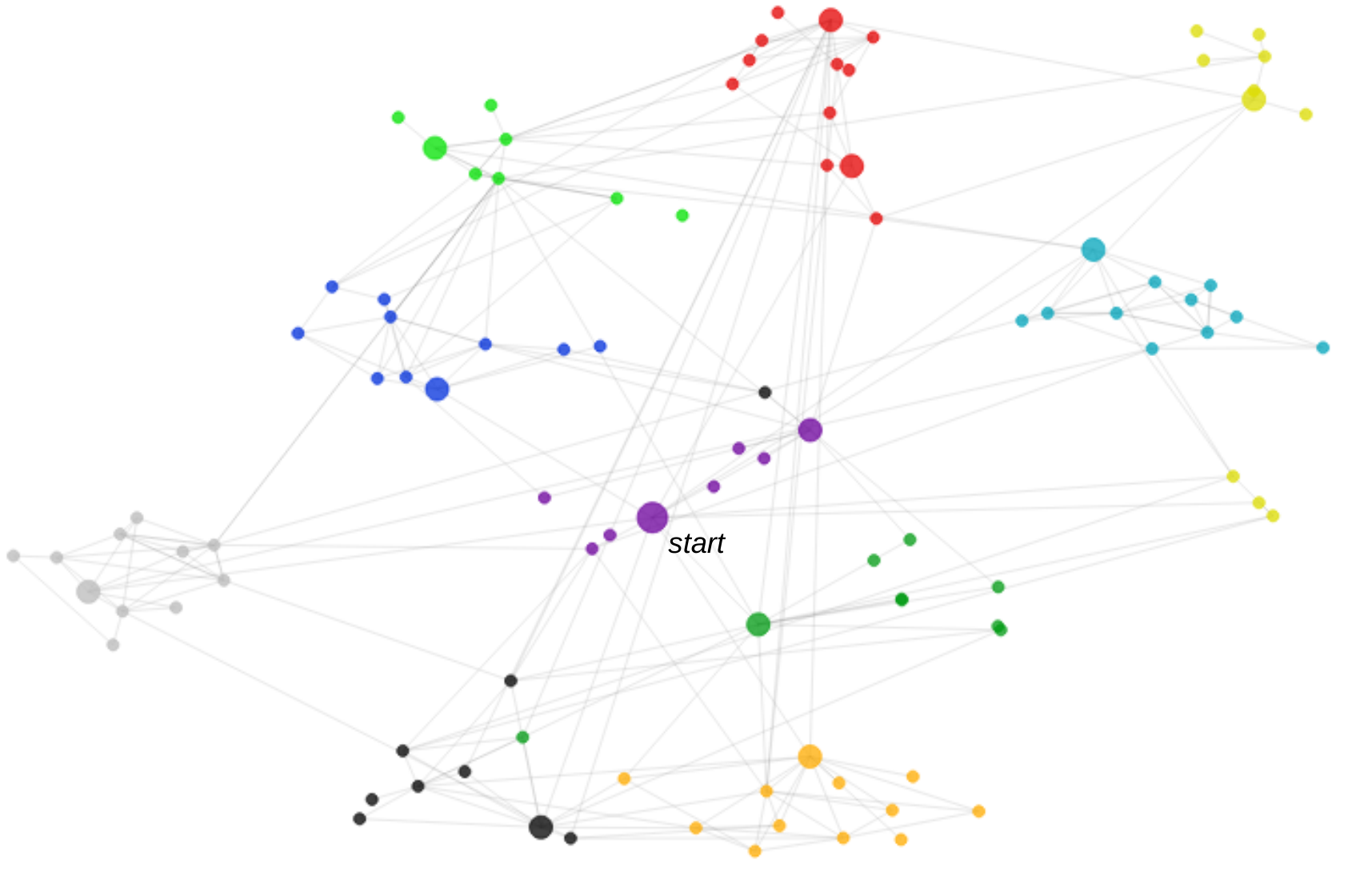} & \hspace{-5mm}
\includegraphics[height=4.5cm]{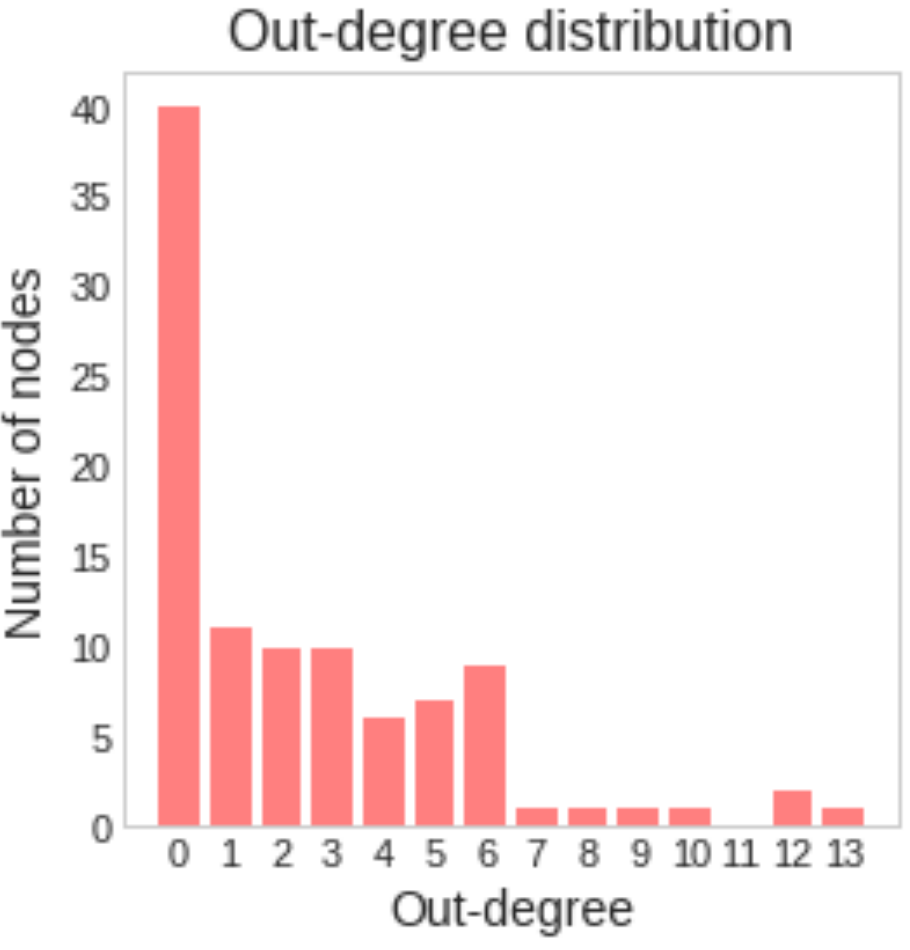}
\end{tabular}
\caption{\textbf{Left:} the constructed graph on 100 vectors from the MNIST8x8 dataset. The optimization is performed over a complete graph. Colors correspond to the MNIST class labels. The nodes providing efficient graph navigation (hubs) are denoted by large sizes. Each MNIST class contains up to two hubs. \textbf{Right:} the outdegree histogram for the obtained graph. Most vertices have zero outdegree and only few with degrees greater than six. All high outdegree nodes correspond to hubs. }
\label{fig:toy}
\end{figure*}


\subsection{Training on large databases}
For large-scale problems, our approach becomes limited by the number of edges it can consider. Namely, if the agent is allowed to draw edges between arbitrary vertices, the number of edges grows quadratically with the database size. Hence it is practically infeasible to train such an agent on the complete graph built upon large databases typical for NNS problems. To mitigate this issue, we limit the agent to a predefined subset of edges. Namely, we construct one of the existing heuristics-based graphs and allow our agent to select edges from that graph. In all our experiments, the initial graph vertex degrees are equal or slightly larger than in baseline graphs which, by themselves, appear to have redundant edges. 

To speed-up training, we also employ the following heuristic. If an agent's prediction for a particular edge is overconfident for a long period during training, we consider this edge deterministic and do not optimize over it. This heuristic reduces optimization problem complexity and allows the agent to concentrate on adjusting predictions for more uncertain edges. As a possible research direction, it is interesting to develop an effective method for expanding the search space, e.g. by interactively adding new edges during training.



\section{Experiments}

\label{sect:experiments}

In this section, we evaluate and analyze graphs constructed by our approach. First, we visualize a toy graph, learned for a small dataset, and describe several interesting observations. Then, we provide an experimental comparison of the constructed graphs with state-of-the-art graph-based methods and analyse the emerging properties of the learned graphs.




\subsection{Toy example}\label{sect:toy}

We visualize graphs constructed by our method on a small subset of the MNIST8x8 \cite{mnist} dataset. Namely, we sample $100$ $64$-dimensional vectors for the base set and use the entire dataset as training queries.

In this experiment, we use greedy search as the search algorithm: we choose the next vertex as the closest one among neighbors of the current query position. The RL agent starts training from a complete graph, and we set $DCS_{max}{=}150$. After the training we manually remove edges that are never used by the search algorithm. Such edges affect neither recall nor $DCS$ and only bring noise to degree distribution.

At convergence, the constructed graph achieves $0.957$ recall. On average, the search algorithm requires $22$ $DCS$ and terminates after $2.85$ graph hops. The average outdegree is reduced from $99$ to $2.45$.

Finally, we project the base vectors onto 2D plane, using tSNE \cite{maaten2008visualizing} and illustrate the graph structure on \fig{toy} (left). The vertex colors correspond to the MNIST class labels. The $start$ vertex is the entry point for the search algorithm --- a medoid of the base set.

In order to analyze the properties of the learned similarity graph, we run the search algorithm for all queries and aggregate the following statistics: (1) how often each node is visited and (2) for what number of queries each node is an actual nearest neighbor. Below we highlight several observations from \fig{toy} and explain our intuition about graphs appropriate for the NNS problem.

\begin{itemize}
    \item We observe an appearance of few nodes, so-called \textit{hubs}, that provide efficient navigation over the graph. Each MNIST class contains one or two hubs. The $start$ node is connected to hubs for fast navigation to a query region. At the first step, the search navigates to one of the hubs. Then, it either finds the answer or transits to another local hub, which is closer to an actual nearest neighbor. The existence of hubs allows the search algorithm to reach answers just in two or three hops. At the same time, the average node outdegree is low, as the number of hubs is small.

    \item Most vertices do not participate in graph navigation. The search algorithm mostly visits such a vertex if it is the actual nearest neighbor for a given query. These vertices are usually terminal, hence their outdegrees are almost zeros.
    
    
\end{itemize}

\begin{figure*}
\noindent
\centering

\begin{tabular}{cc}
SIFT100K & DEEP100K \\\hspace{-1mm}
\includegraphics[height=4.5cm]{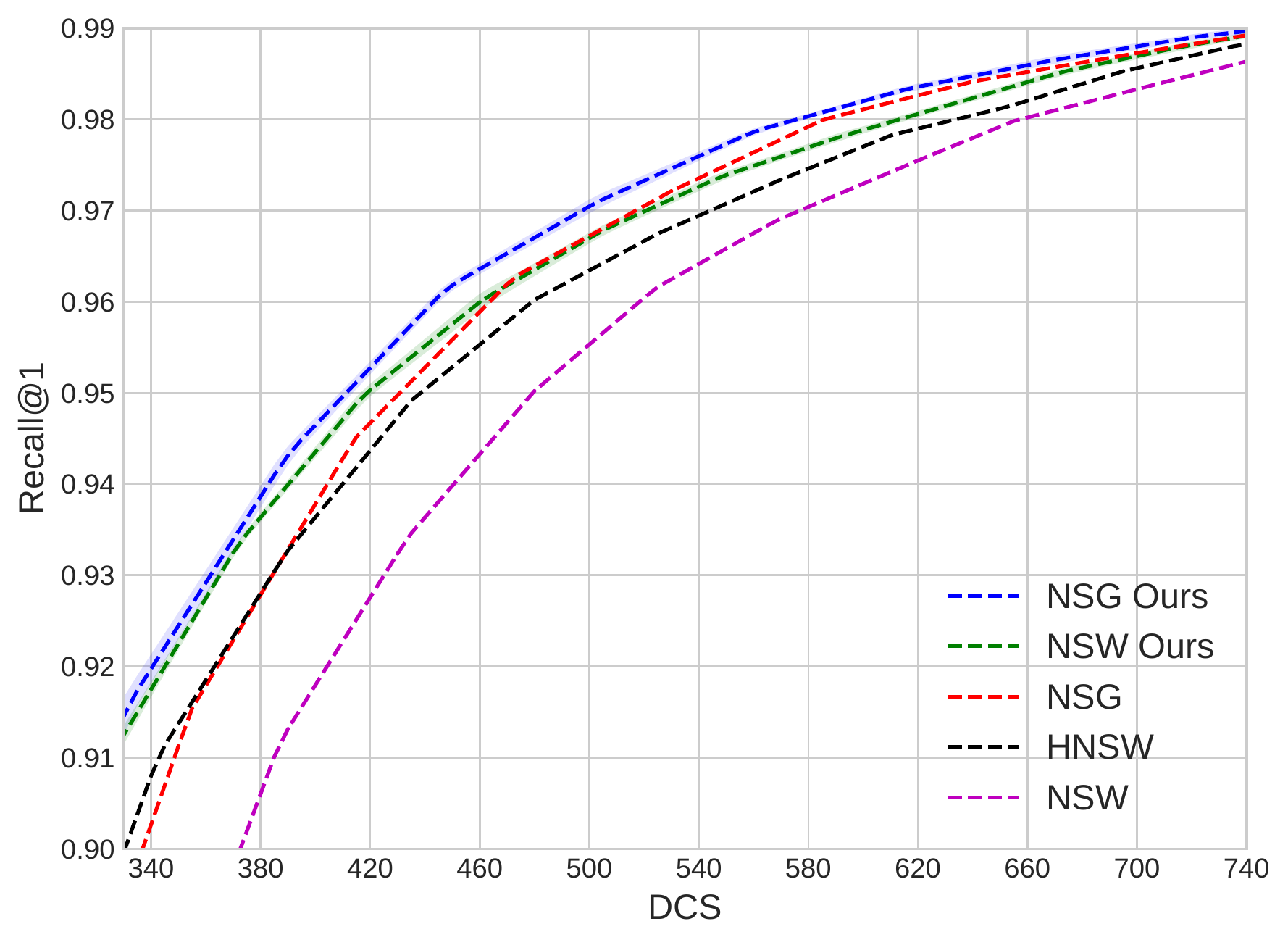} & 
\includegraphics[height=4.5cm]{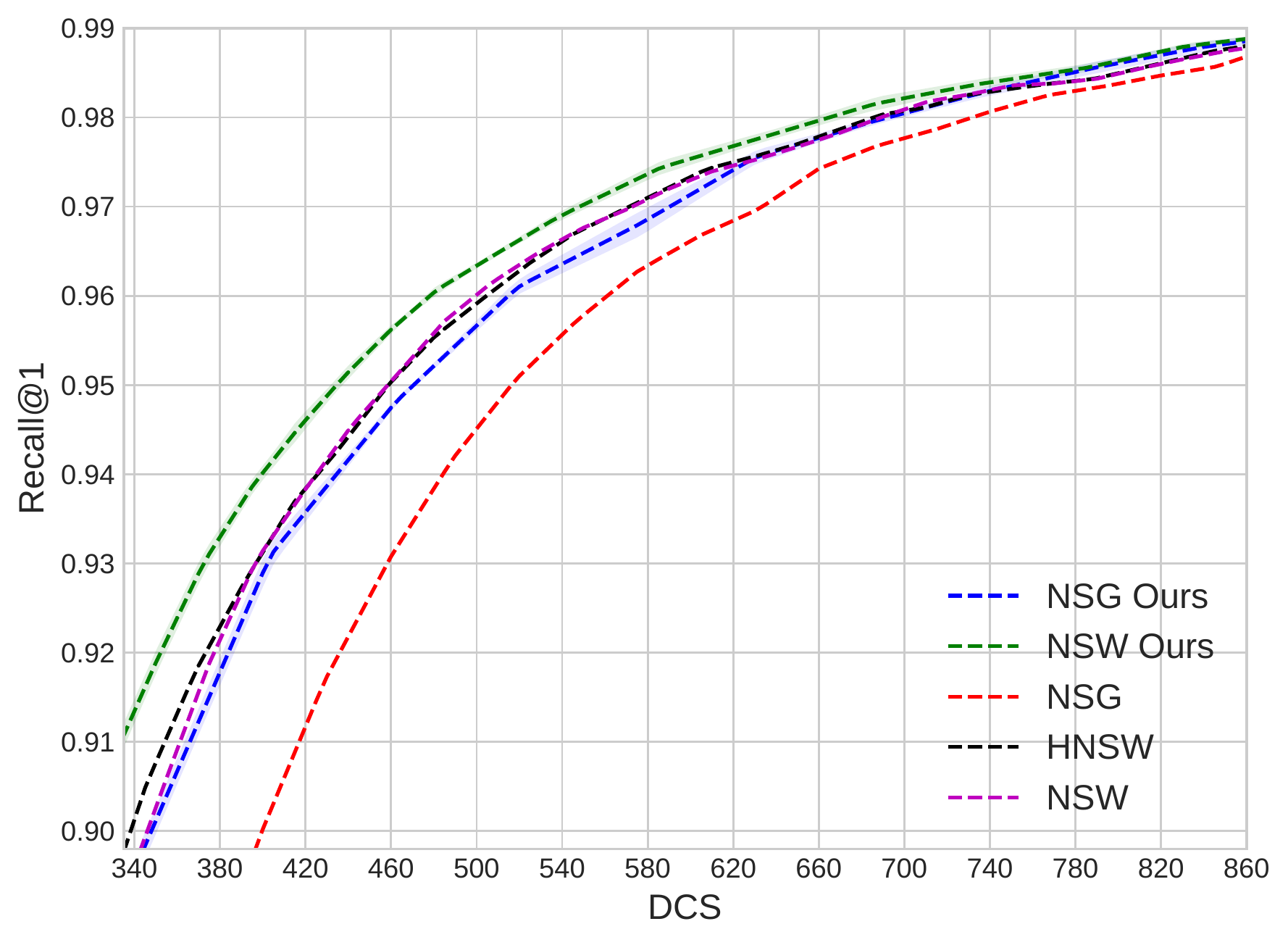} \hspace{-6mm}\\
\end{tabular}

\vspace{-4mm}
\caption{$Recall@1$ values as functions of distance computations $DCS$ on the SIFT100K and DEEP100K datasets.\vspace{-3px}}
\label{fig:sift}
\end{figure*}

\begin{figure*}
\noindent
\centering

\begin{tabular}{ccc}
SIFT1M & DEEP1M & GloVe1M \\ \hspace{-3mm}
\includegraphics[height=4.6cm]{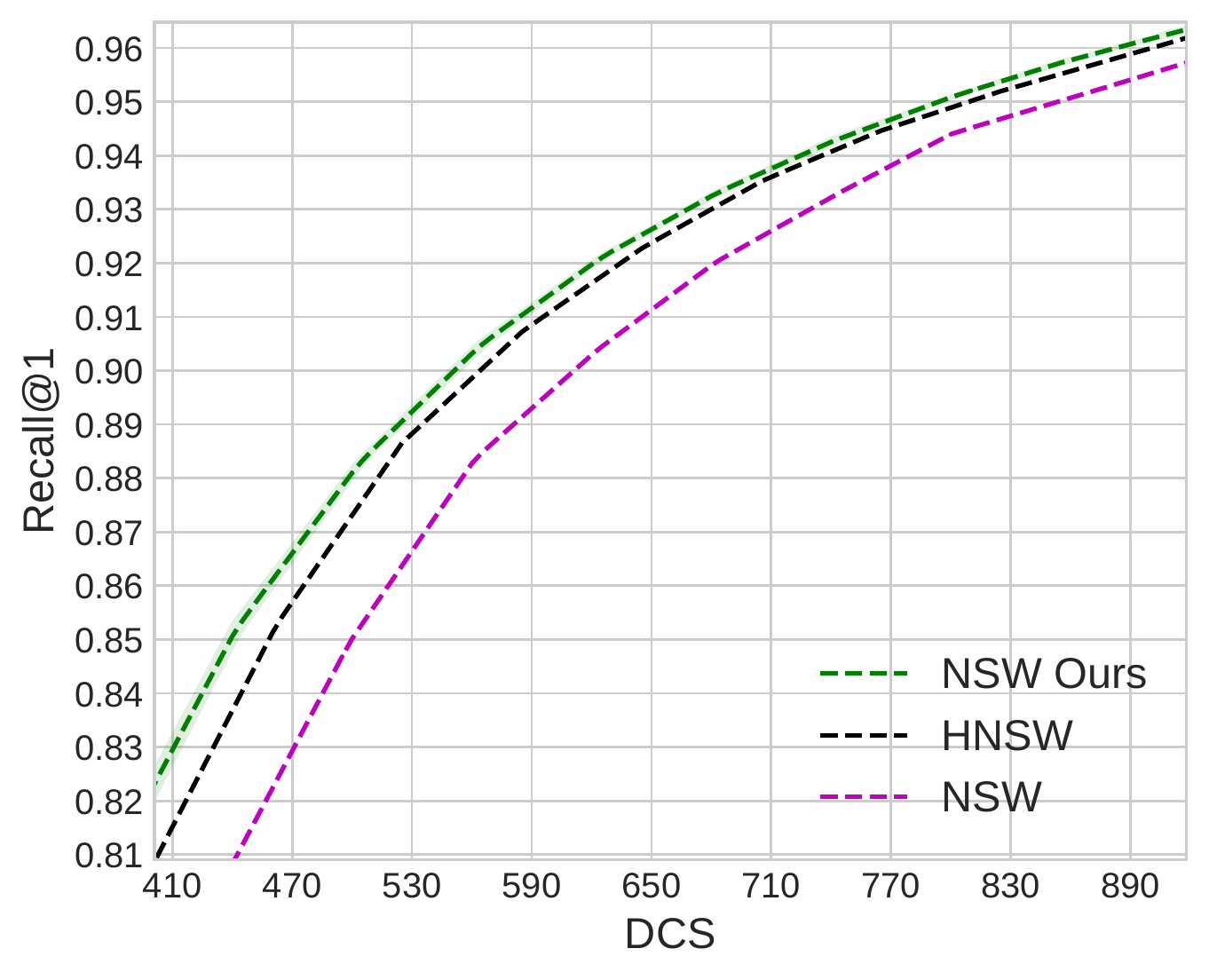} & \hspace{-4mm}
\includegraphics[height=4.6cm]{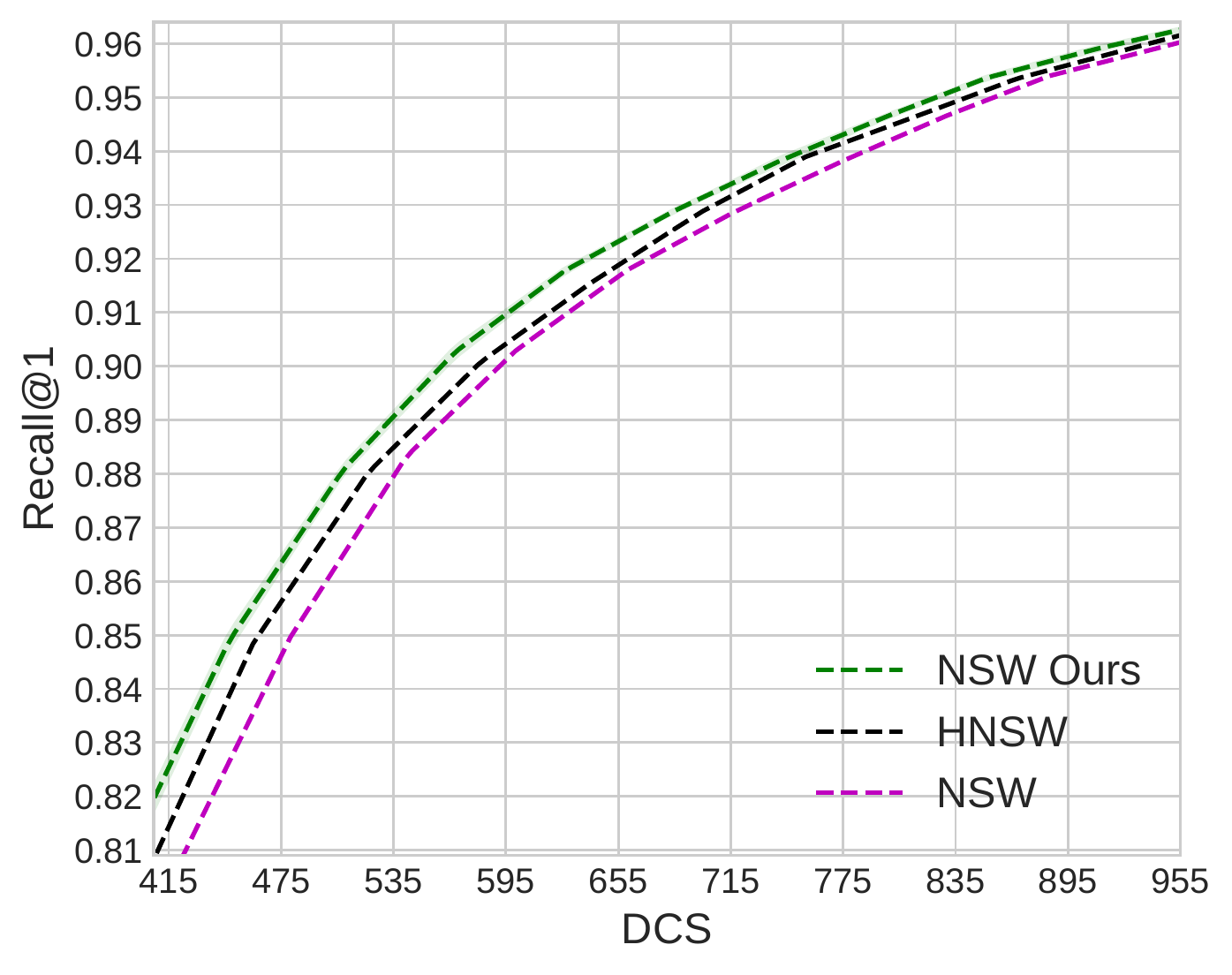} & \hspace{-4mm} 
\includegraphics[height=4.6cm]{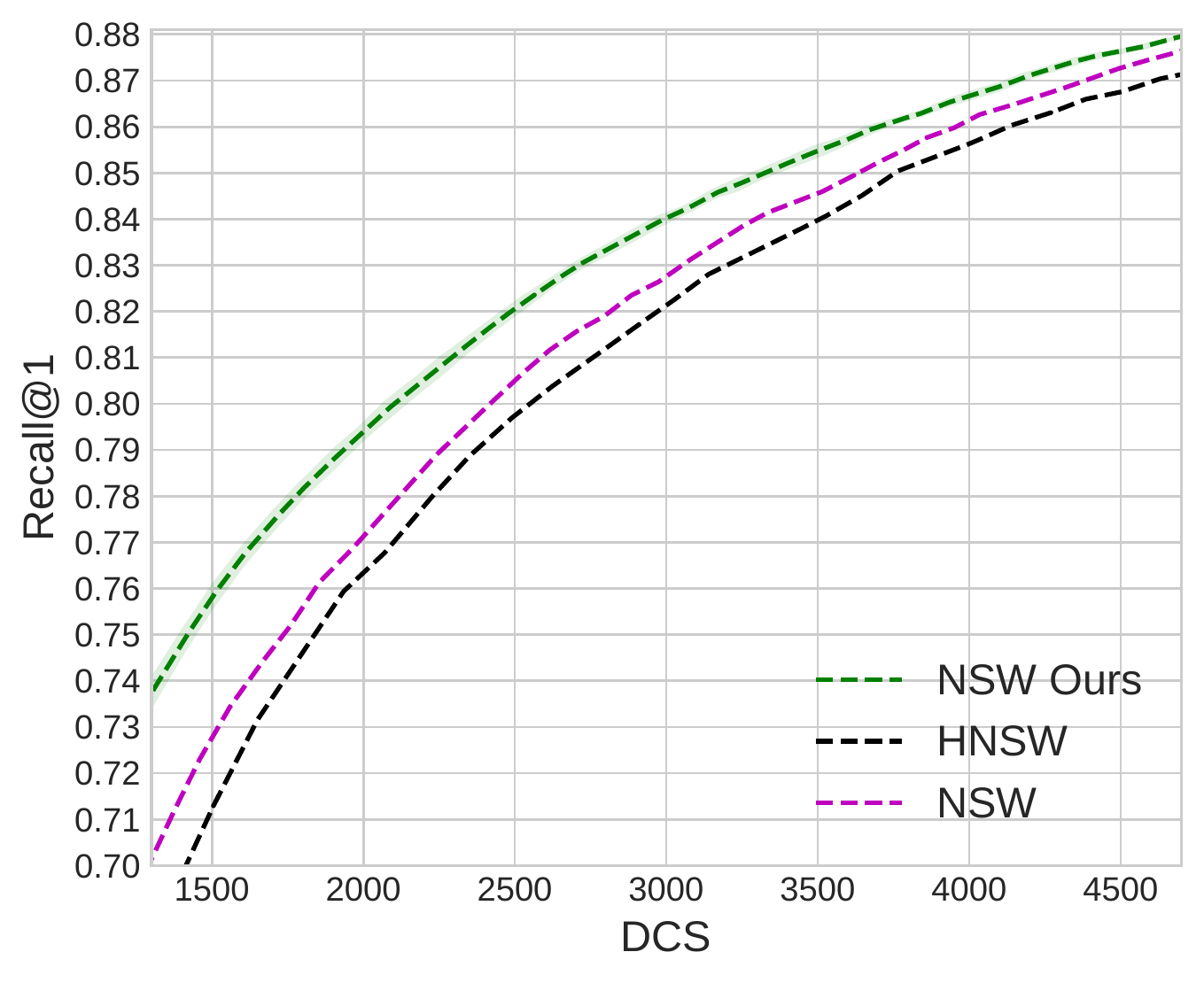} \hspace{-4mm}\\
\end{tabular}

\vspace{-4mm}
\caption{$Recall@1$ values as functions of distance computations $DCS$ on the SIFT1M, DEEP1M and GloVe1M datasets.\vspace{-3px}}
\label{fig:million}
\end{figure*}

Additionally, we plot the outdegree histogram for the constructed graph on \fig{toy} (right). Most vertices have zero outdegrees and only few have a degree greater than six. This roughly resembles the truncated power-law distribution over outdegrees. Interestingly, all high-outdegree nodes are hubs. A prior work\cite{malkov2016growing} investigates the properties of graphs with truncated power-law degree distribution for the NNS problem and shows that such degree distribution is likely to provide an efficient search. In our approach, such properties emerge naturally from search performance optimization over the complete graph.

\begin{figure*}[h]
    \centering
    \begin{tabular}{cc}
    \hspace{-3mm}\includegraphics[width=242px]{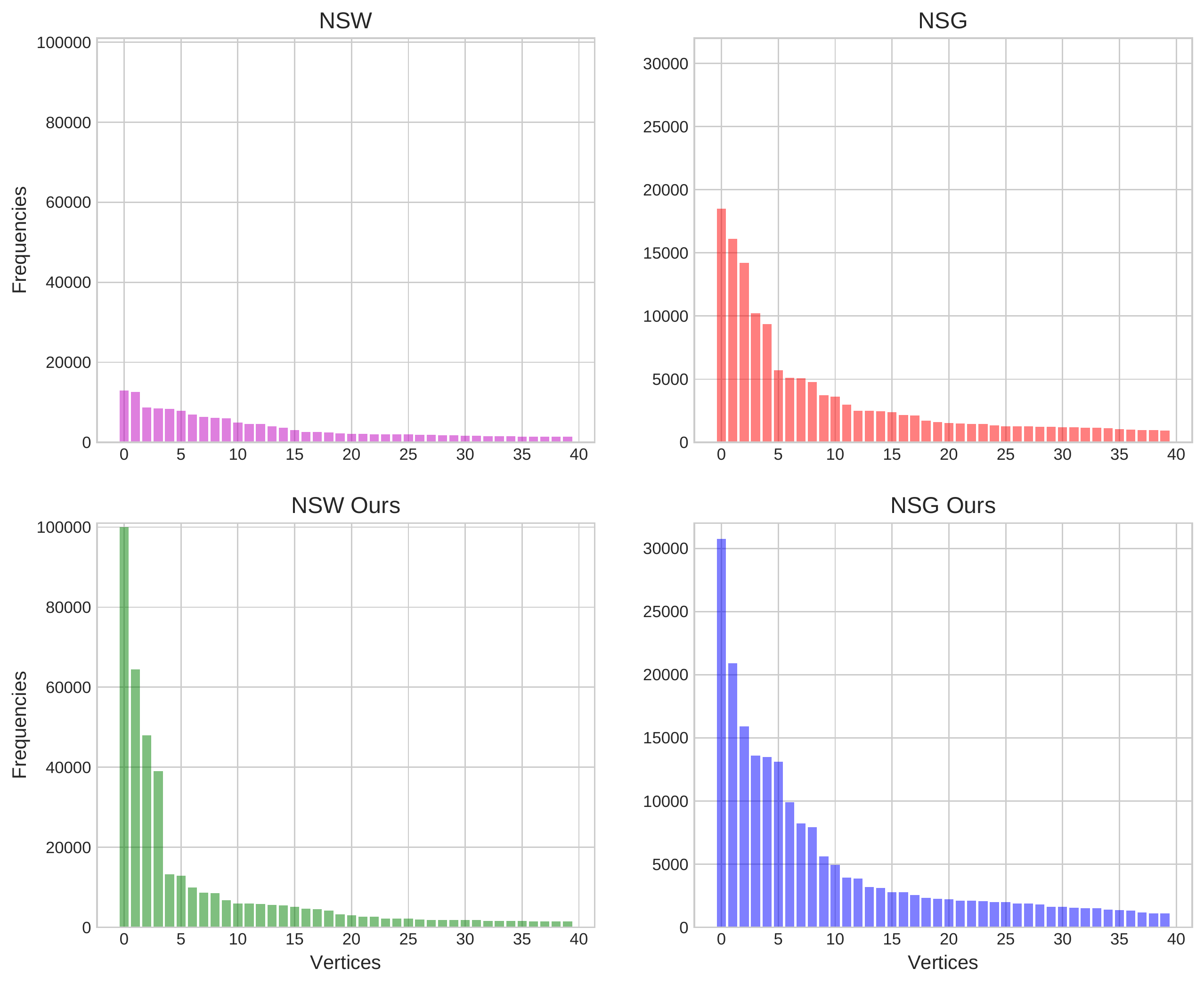} &  \hspace{-3mm}\includegraphics[width=242px]{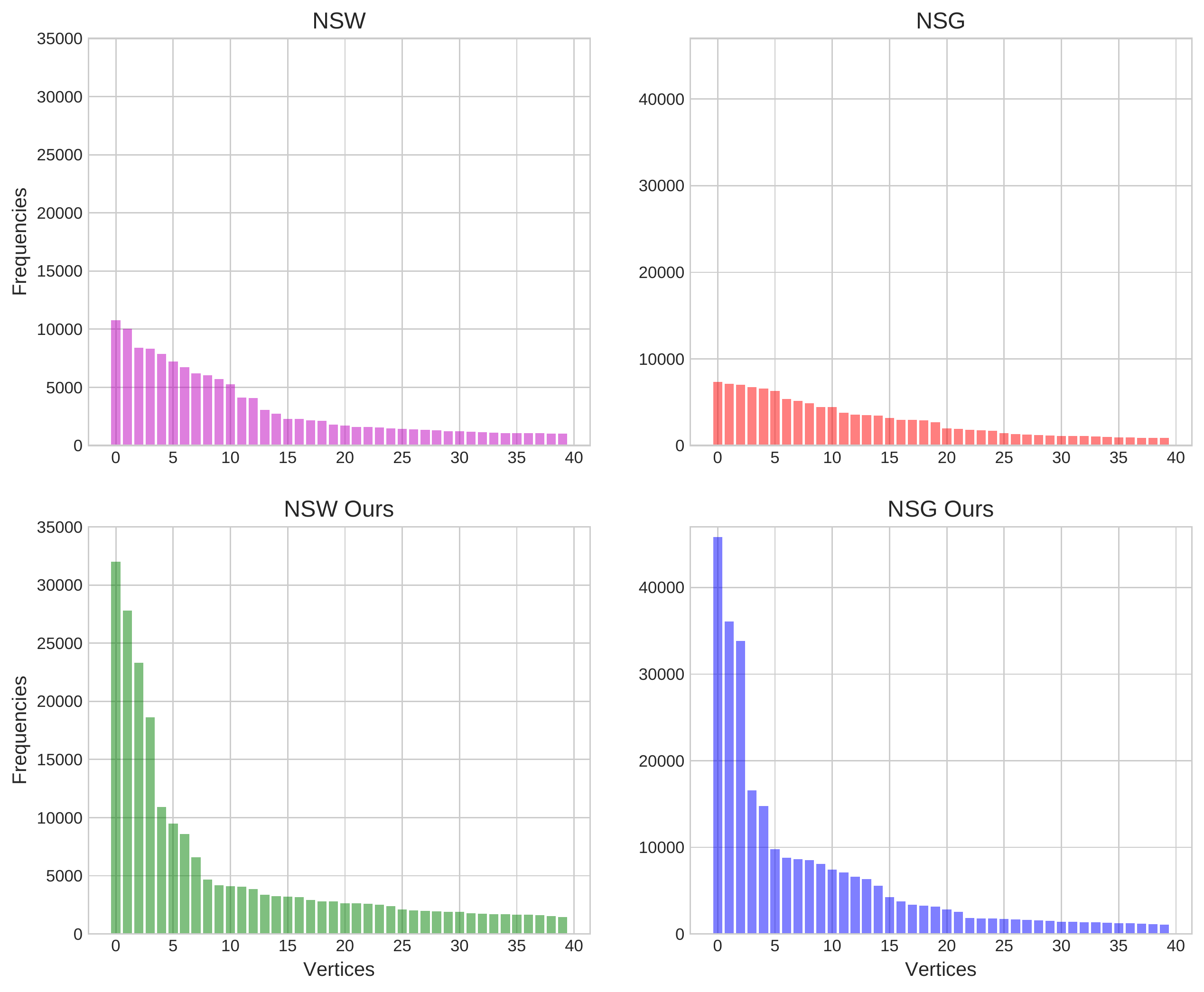}
    \\
     SIFT100K & DEEP100K 
    \end{tabular}
    \caption{Search visitation frequencies for 40 most visited vertices, sorted by frequency (except for \textit{start} vertex). The top row represents the baseline graphs; the bottom row depicts their counterparts optimized by our method.}
    \label{fig:freq}
\end{figure*}

\subsection{Datasets}
We evaluate the proposed approach on three publicly available datasets described below:

\begin{enumerate}
    \vspace{-2mm}
    \item SIFT100K dataset \cite{Jegou11a} is sampled from one million $128$-dimensional SIFT descriptors. We consider 100,000 learn vectors and remained base vectors as train queries. Note, the original learn set contains test queries, therefore we manually remove them. We take 20,000 datapoints for validation. The hold-out 10,000 query vectors are used for evaluation.
    \item SIFT1M dataset contains one million SIFT descriptors sampled from SIFT1B \cite{Jegou11a}. We sample one million train queries from the learn set. Again, we leave 20,000 queries for validation and evaluate on original 10,000 hold-out queries. 
    \item DEEP100K dataset \cite{BabenkoCVPR16} is a subset of one billion of $96$-dimensional CNN-produced feature vectors of natural images from the Web. The base set contains 100,000 vectors. We sample 200,000 train and 20,000 validation queries from the learn set. For evaluation, we use the original 10,000 queries.
    \item DEEP1M dataset is the same as DEEP100K where the base and learn sets are extended to one million datapoints.
    \item GloVe1M dataset \cite{pennington2014glove} contains 2.2 millions of 300-dimensional word embeddings trained on Common Crawl. We split them on one million base set, one million learn set, 20,000 queries for validation and 10,000 queries for evaluation.   
\end{enumerate}

\subsection{Search performance evaluation}\label{sect:evaluation}

Here we compare the graphs constructed with our method to state-of-the-art baselines on the SIFT100K and DEEP100K datasets. Namely, we evaluate:

\begin{itemize}
\setlength\itemsep{0.4em}
    \item \textbf{HNSW}: one of the current state-of-the-art graphs proposed in \cite{HNSW}; this approach exploits the nested hierarchy of navigable small-world graphs constructed on the database subsets to obtain a start vertex.
    \item \textbf{NSW}: the bottom layer of HNSW graph. The search starts from the fixed vertex for all queries.
    \item \textbf{NSG}: another state-of-the-art similarity graph method \cite{NSG}; NSG does not use any additional indexing structure and starts the search from the database medoid.
    \item \textbf{NSW Ours}: RL approach applied to the NSW graph. 
    \item \textbf{NSG Ours}: RL approach applied to the NSG graph. 
\end{itemize}

We tune hyperparameters for all baseline graphs in each recall region. All parameters for the graphs listed above are reported in the supplementary materials. Note that the proposed RL-based approach can also be applied to graphs with additional indexing structures (e.g., HNSW, NGT). However, we leave it beyond the scope of our evaluation. 

As a primary performance measure, we use $Recall@1$, which is calculated as a rate of queries for which the search algorithm successfully finds the actual nearest neighbor.

Most million-scale experiments converge within ${\sim}24$ hours on a single GPU GeForce 1080Ti. We rerun the RL approach at least five times for each graph and draw its mean and standard deviation. The plots for the SIFT100K and DEEP100K datasets are presented on \fig{sift}, and the plots for SIFT1M, DEEP1M, GloVe1M are presented on \fig{million}.

For all datasets, we observe a consistent improvement over corresponding baseline graphs. We highlight several key observations below:


\begin{itemize}
\setlength\itemsep{0.0em}
    \item  On SIFT100K, the optimized NSG consistently outperforms all other evaluated graphs. In particular, we observe up to ${\sim}1\%$ improvement compared to the top-performing NSG baseline. On DEEP100K, the optimized NSW graph also outperforms HNSW/NSW graph by up to ${\sim}1\%$. For $99{+}\%$ $Recall@1$ region, the gains become insignificant. Note that NSG graphs are superior on SIFT data, while NSW/HNSW performs better on the DEEP100K dataset. This is a weakness of heuristic-based similarity graphs: different heuristics are more appropriate for different data. Our RL-based approach may significantly reduce the gap in performance. E.g., while NSG outperforms NSW by up to ${\sim}2.5\%$ on SIFT100K, the maximum gap between optimized graphs reduces to ${\sim}0.4\%$. On DEEP100K, NSW/HNSW outperforms NSG by up to ${\sim}3.0\%$, while, for NSW Ours and NSG Ours, the maximum difference is ${\sim}1.3\%$.
    
    \item On all datasets, we observe more significant gains for lower $Recall@1$ regions. While our hypothesis that the RL approach mainly influences the navigation properties of similarity graphs, this observation is consistent with the fact that navigation properties lose their value if the search algorithm's heap size increases. 
    
    \item On all datasets and all $Recall@1$ regions, the optimized NSW is superior or equal to HNSW, which exploits an additional indexing structure for better navigation. Therefore, the nested hierarchy of graphs is redundant and can be replaced by its bottom layer with improved navigation properties.
    
    \item On the most challenging dataset, GloVe1M, NSW/HNSW graphs demonstrate much worse performance due to the high intrinsic dimensionality of the word embeddings. For this dataset, our approach mitigates the issues of NSW/HNSW graphs and outperforms baselines by ${\sim}0.4\%$ at $88\%$ $Recall@1$ point.

\end{itemize}

\subsection{Graph properties analysis} 

In this section, we analyze the emerging properties of graphs learned by the proposed algorithm. Our primary hypothesis is that the advantage of our method in terms of search efficiency is attributed to its ability to learn more specialized roles for graph vertices, similarly to what we observed in the toy experiment.

In order to test this hypothesis, we study the statistical properties of frequently visited vertices. In both NSW and NSG graphs, there is a small subset of vertices that help the search procedure to navigate during the first few graph hops. Hence, an improvement in these vertices may have a substantial effect on the overall search efficiency.


We consider 40 vertices that are the most frequently visited by the search algorithm. For each vertex, we count its number of visits over $10^5$ training queries. The obtained numbers of visits for baseline graphs and graphs produced by our method are presented on \fig{freq}. 

\fig{freq} clearly indicates that graphs produced by our method have a more peaky distribution over vertex visit frequencies compared to both baselines. In other words, directly optimizing graph for nearest neighbor search produces more specialized navigation vertices.

Interestingly, our RL approach can also learn a new starting vertex, see NSW Ours on SIFT100K in \fig{freq}. The agent omits all edges in initial starting vertex except one. Hence, for every query the search procedure goes to the new starting node by performing only one distance computation. Note that ``peakyness'' of the distributions from \fig{freq} correlates with relative performance of heuristics-based graphs on different datasets. For instance, on SIFT100K NSG has more pronounced hubs and outperforms NSW on this dataset, see \fig{sift}. In contrast, on DEEP100K, NSW has more ``peaky'' distribution compared to NSG and provides superior search performance.

We conjecture that our algorithm is better able to learn the edges for the navigation vertices, achieving more accurate routing, compared to heuristics-based counterparts. 




\subsection{Comparison to heuristic methods}
In this experiment, we evaluate our approach against one of the heuristic methods, which can be used for similarity graph improvement. 

Here, we consider \textit{magnitude-based pruning}, where the weights for each edge are computed as follows: 
\begin{equation}
w_{ij} = \frac{n\_visited\_e_{ij} + \lambda}{n\_visited\_v_{i} + \lambda \cdot outdegree(v_i)}
\end{equation}
, where $n\_visited\_e_{ij}$ and $n\_visited\_v_{i}$ correspond to visitation frequencies for edge $e_{ij}$ and vertex $v_i$ respectively. We compute these frequencies by running search procedure on training queries. The only hyperparameter $\lambda$ plays a smoothing role, discouraging radical pruning of rarely visited vertices. In our experiments we always use $\lambda{=}0.1$. Then, we tune a weight threshold to maximize performance for validation queries. Finally, all edges whose weights are below the threshold are pruned. 

\begin{figure}
\noindent
\centering
\includegraphics[height=6.2cm]{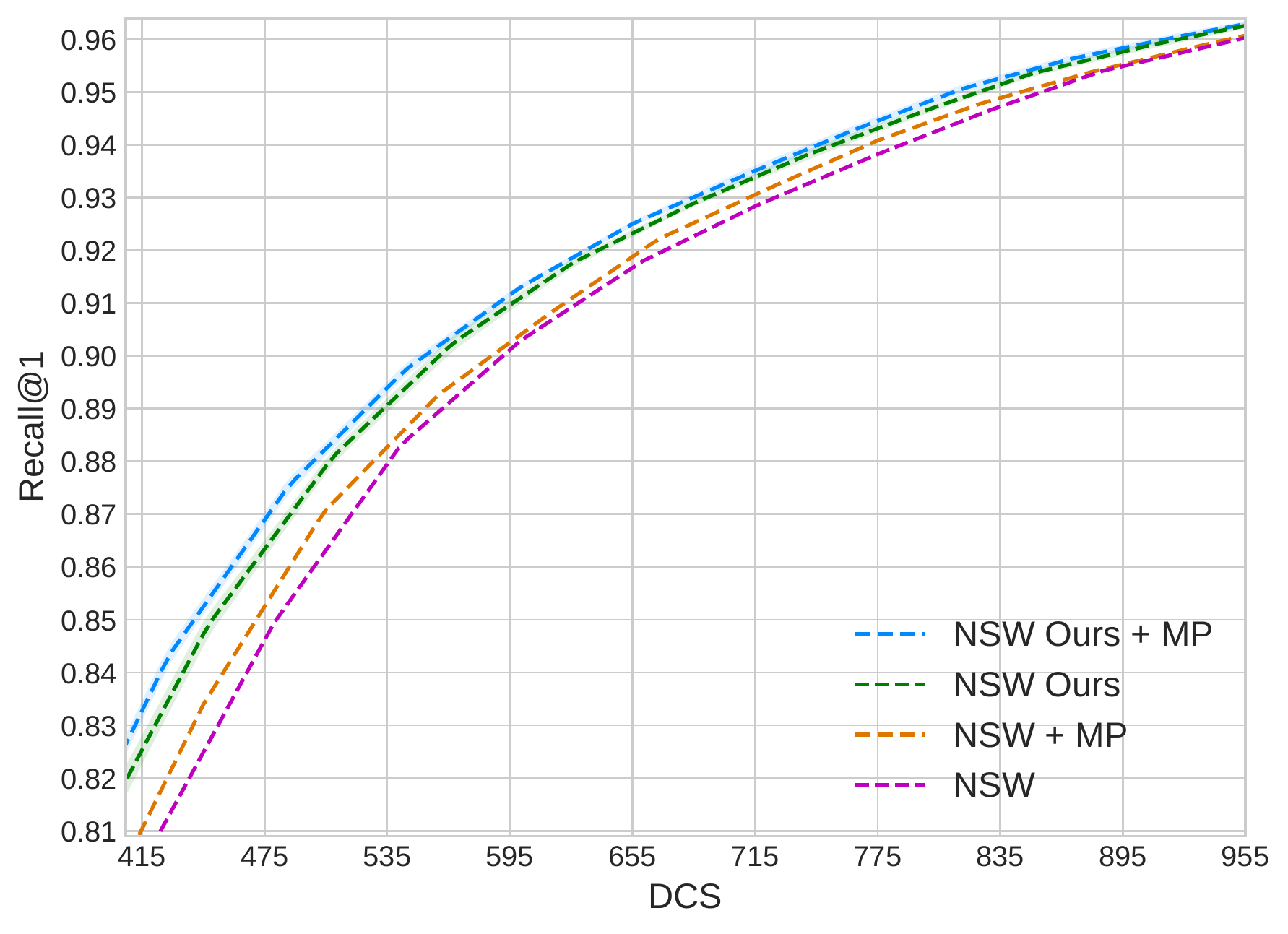}
\vspace{-8px}
\caption{Comparison to magnitude-based pruning (MP) for the NSW graph on DEEP1M dataset.}
\label{fig:mp}
\end{figure}

We compare our RL-based approach and magnitude pruning applied to the NSW graph on DEEP1M, see \fig{mp}. Our method outperforms magnitude pruning across all distance computation budgets. Finally, we apply magnitude pruning to the graph constructed by RL and observe that it also slightly improves the performance.


\section{Conclusion}
\label{sect:conclusion}

In this paper, we introduce a new algorithm for similarity graph construction that explicitly optimizes an adjacency matrix, maximizing the search quality for a large set of training queries. The algorithm defines a probabilistic model of the graph in terms of its edge probabilities and then learns these probabilities in a reinforcement learning scenario. We show that the proposed approach allows to improve the performance of similarity graphs constructed by heuristics.


\bibliography{example_paper}
\bibliographystyle{icml2020}

\newpage~\newpage~

\vspace{-9mm}

\subsection*{\centering A.1 Hyperparameters SIFT100K}

\begin{table}[!htp]
\hspace{-8mm}
\addtolength{\tabcolsep}{-2pt}
\renewcommand\arraystretch{1.2}
\begin{tabular}{|c|c|c|c|c|c|}
\hline
& NSW & HNSW & NSG & NSW Ours & NSG Ours\\
\hline
$M$ & 12 & 12 & - & 12 & -\\
$ef_{construction}$ & 500 & 300 & - & 300 & -\\
\hline
$R$ & - & - & 24 & - & 24 \\
$K$ & - & - & 200  & - & 200\\
\hline
$ef_{search}$ & - & - & - & 10 & 10\\
$DCS_{max}$ & - & - & - & 1200 & 1500\\
$C_{entropy}$ & - & - & - & 0.01 & 0.001\\
\hline
\end{tabular}
\end{table}

\subsection*{\centering A.2 Hyperparameters SIFT1M}

\begin{table}[h!]
\centering
\addtolength{\tabcolsep}{-2pt}
\renewcommand\arraystretch{1.2}
\begin{tabular}{|c|c|c|c|}
\hline
& NSW & HNSW & NSW Ours \\
\hline
$M$ & 14 & 14 & 14\\
$ef_{construction}$ & 500 & 500 & 500\\
\hline
$ef_{search}$ & - & - & 12\\
$DCS_{max}$ & - & - & 1500\\
$C_{entropy}$ & - & - & 0.01\\ 
\hline
\end{tabular}
\end{table}

\newpage

\subsection*{\centering A.3 Hyperparameters DEEP100K}

\begin{table}[!htp]
\hspace{-2mm}
\addtolength{\tabcolsep}{-2pt}
\renewcommand\arraystretch{1.2}
\begin{tabular}{|c|c|c|c|c|c|}
\hline
& NSW & HNSW & NSG & NSW Ours & NSG Ours\\
\hline
$M$ & 12 & 12 & - & 12 & -\\
$ef_{construction}$ & 300 & 300 & - & 300 & -\\
\hline
$R$ & - & - & 24 & - & 24\\
$K$ & - & - & 200  & - & 200\\
\hline
$ef_{search}$ & - & - & - & 10 & 10\\
$DCS_{max}$ & - & - & - & 1000 & 1500\\
$C_{entropy}$ & - & - & - & 0.01 & 0.001 \\
\hline
\end{tabular}

\end{table}

\subsection*{\centering A.4 Hyperparameters DEEP1M}

\begin{table}[h!]
\centering
\addtolength{\tabcolsep}{-2pt}
\renewcommand\arraystretch{1.2}
\begin{tabular}{|c|c|c|c|}
\hline
& NSW & HNSW & NSW Ours \\
\hline
$M$ & 14 & 14 & 14\\
$ef_{construction}$ & 500 & 500 & 500\\
\hline
$ef_{search}$ & - & - & 12\\
$DCS_{max}$ & - & - & 1500\\
$C_{entropy}$ & - & - & 0.01\\ 
\hline
\end{tabular}
\end{table}

\subsection*{\centering A.5 Hyperparameters GloVe1M}

\begin{table}[h!]
\centering
\addtolength{\tabcolsep}{-2pt}
\renewcommand\arraystretch{1.2}
\begin{tabular}{|c|c|c|c|}
\hline
& NSW & HNSW & NSW Ours \\
\hline
$M$ & 20 & 28 & 20\\
$ef_{construction}$ & 2000 & 2000 & 2000\\
\hline
$ef_{search}$ & - & - & 5\\
$DCS_{max}$ & - & - & 1000\\
$C_{entropy}$ & - & - & 0.01\\ 
\hline
\end{tabular}
\end{table}




\end{document}